\documentclass[journal]{IEEEtran}

\IEEEoverridecommandlockouts
\usepackage{cite}
\usepackage{amsmath,amssymb,amsfonts,mathtools}
\usepackage[linesnumbered,ruled,vlined]{algorithm2e}
\usepackage{graphicx}
\usepackage{color}
\usepackage{textcomp}
\usepackage{xcolor}
\usepackage{tensor}
\usepackage{float}
\usepackage{nomencl}
\makenomenclature
\usepackage{caption}
\usepackage{subcaption}
\usepackage[OT1]{fontenc}
\usepackage{multirow}
\usepackage{gensymb}
\usepackage{hyperref}
\usepackage{bm}
\usepackage{booktabs}

\newtheorem{myThe}{Theorem}

\graphicspath{{./figs/}}
\DeclareGraphicsExtensions{.png,.jpg,.eps,.pdf}

\title{Distributed Pose-graph Optimization with Multi-level Partitioning for Collaborative SLAM}
\author{Cunhao Li, Guanghui Guo, Peng Yi, Yiguang Hong
\thanks{The paper was sponsored by the National Key Research and Development Program of China under No.2022YFA1004701, the National Natural Science Foundation of China under No.72271187 and No.62373283, and partially by Shanghai Municipal Science and Technology Major Project No.2021SHZDZX0100, and National Natural Science Foundation of China (Grant No.62088101).

Cunhao Li and Guanghui Guo are with the Department of Control Science and Engineering, Tongji University,Shanghai.

Peng Yi and Yiguang Hong are with the Department of Control Science and Engineering \& National Key Laboratory of Autonomous Intelligent Unmanned Systems \& Frontiers Science Center for Intelligent Autonomous Systems, Ministry of Education, Tongji University, Shanghai,China(e-mail: licunhao@tongji.edu.cn;18916106931@163.com yipeng@tongji.edu.cn; yghong@tongji.edu.cn).
}
}

\begin{document}
  \maketitle
  \pagestyle{empty}  
  \thispagestyle{empty} 
\begin{abstract}

The back-end module of \textit{Distributed Collaborative Simultaneous Localization and Mapping} (DCSLAM) requires solving a nonlinear \textit{Pose Graph Optimization} (PGO) under a distributed setting, also known as $SE(d)$-synchronization.
Most existing distributed graph optimization algorithms employ a simple sequential partitioning scheme, which may result in unbalanced subgraph dimensions due to the different geographic locations of each robot, and hence imposes extra communication load.
Moreover, the performance of current Riemannian optimization algorithms can be further accelerated.
In this letter, we propose a novel distributed pose graph optimization algorithm combining multi-level partitioning with an accelerated Riemannian optimization method.
Firstly, we employ the multi-level graph partitioning algorithm to preprocess the naive pose graph to formulate a balanced optimization problem.
In addition, inspired by the accelerated coordinate descent method, we devise an \textit{Improved Riemannian Block Coordinate Descent} (IRBCD) algorithm and the critical point obtained is globally optimal.
Finally,  we evaluate the effects of four common graph partitioning approaches on the correlation of the inter-subgraphs, and discover that the \texttt{Highest} scheme has the best partitioning performance.
Also, we implement simulations to quantitatively demonstrate that our proposed algorithm outperforms the state-of-the-art distributed pose graph optimization protocols\footnote{Our code is publicly available: https://github.com/tjcunhao/dpo.}.
\end{abstract}


\section{Introduction}\label{sect:intro}
In recent years, multi-robot systems have been widely applied in many domains such as field search and rescue \cite{tian2020search}, area coverage \cite{kim2015active} and environment exploration \cite{mahdoui2020communicating} due to their efficient collaboration, robust generalization, and flexible reconfiguration.
For these complex tasks, \textit{collaborative localization and mapping} (CSLAM) is a critical technology that enables each robot to cooperatively estimate its poses and build a global map in a large-scale unknown environment, thus enhancing the autonomy and intelligence of the multi-robot system.
According to whether there is a central computing node, CSLAM systems can be classified into two types: centralized and distributed \cite{cadena2016past}. Distributed CSLAM only depends on the local perception, computation, and communication of peer robots, which makes it more scalable \cite{chung2018survey}.

\begin{figure}[t]
	\centering
	\begin{subfigure}{0.45\linewidth}
		\centering
		\includegraphics[width=1.0\linewidth]{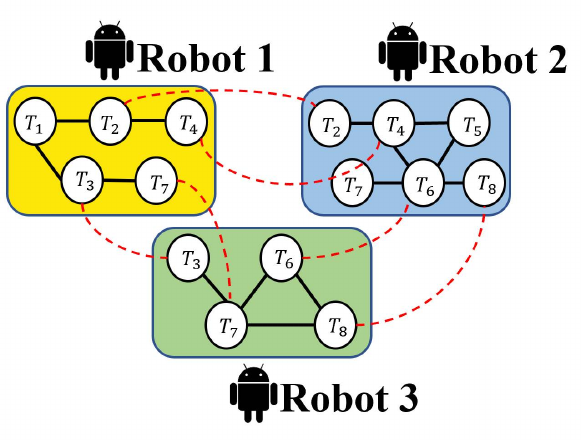}
		\caption{Sequential Partitioning}
		\label{se}
	\end{subfigure}
	\centering
	\begin{subfigure}{0.45\linewidth}
		\centering
		\includegraphics[width=0.96\linewidth]{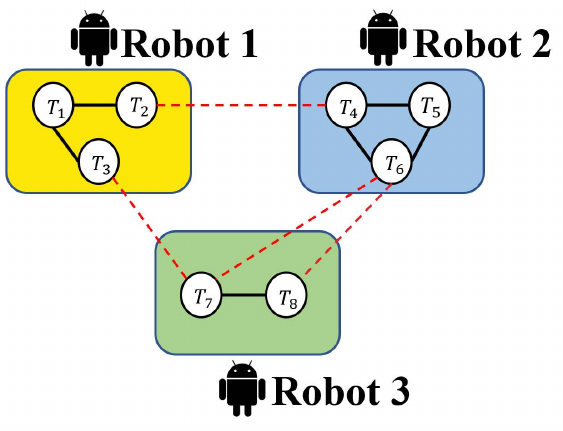}
		\caption{Graph Partitioning}
		\label{graph}
	\end{subfigure}
	\captionsetup{font={small}}
    \caption{Representation of (a) sequential partitioning and (b) graph partitioning.
     We denote the pose variable blocks by $T_1, \cdots,T_8$, use solid black lines to represent the loop closure within the robot, and use dashed red lines to represent the relative pose relationship between the robots.
     (a) and (b) exhibit the subgraphs obtained by each robot after sequential partitioning and graph partitioning, respectively.}
	\label{com}
\end{figure}

\textit{Distributed Pose Graph Optimization} (DPGO) is a core mathematical model abstracted from the backend module of the distributed CSLAM system.
It involves each robot solving a least squares problem in $SE(d)$ space based on the maximum likelihood estimation of the mismatch between the inter-robot loop closure estimated pose and the single-robot relative measurement with noise.
In this case, each node in the pose graph can be regarded as a block of variables composed of rotation variables and translation variables.
The geodesic distance is used to measure the inconsistency between poses, so this least squares problem belongs to Riemannian manifold optimization \cite{6630774}.

However, there are still some open problems for this large-scale distributed nonlinear optimization problem.
Existing distributed graph optimization algorithms mostly rely on the ownership of data measurement to carry out the naive sequential partitioning on the pose graph, without fully exploiting the characteristics of the graph itself, such as connectivity, sparsity, and coupling degree among variables, as shown in Fig.1(a).
This leads to imbalanced dimensions of the optimization subproblem constructed by each robot, which hence increases the communication overhead and reduces the algorithm efficiency.
Besides, there is still room for further improving the performance of the mainstream distributed Riemannian manifold optimization algorithms.

In order to address the above problems, this letter proposes a novel distributed graph optimization algorithm that integrates multi-level graph partitioning with an acceleration method.
We apply the multi-level graph partitioning algorithm to merge the nodes with high coupling degrees in the original pose graph, then split the compressed graph into several blocks and make appropriate fine-grained adjustments.
The graph partitioning approach enables each robot to obtain a balanced pose subgraph, which provides a good initialization for the subsequent distributed optimization (see Fig.1(b)).

Due to the non-convexity of the distributed PGO problem, we reformulate it as a \textit{Low-Rank Convex Relaxation} (LRCR) form and solve it with the proposed \textit{Improved Riemannian Block Coordinate Descent} (IRBCD).
IRBCD is an extension of the \textit{Accelerated Coordinate Descent Method} (ACDM) proposed by Nesterov \cite{doi:10.1137/100802001}, which does not rely on the Lipschitz constant of the objective function and incorporates a generalized momentum term for acceleration.
This letter proves that the IRBCD algorithm can converge to the first-order critical point with global optimality.

In addition, we compare four types of graph partitioning methods belonging to the \textit{Karlsruhe High-Quality Partitioning} (KaHIP) framework (named \texttt{Strong}, \texttt{Eco}, \texttt{Fast}, and \texttt{Highest}, respectively) \cite{meyerhenke2017parallel}, \cite{sanders2012distributed} on standard datasets.
The experimental results demonstrate that the \texttt{Highest} strategy achieves the best partitioning quality among all the strategies in this CSLAM task.
Meanwhile, compared with the state-of-the-art RBCD and RBCD++ \cite{tian2021distributed}, the IRBCD method enhances the accuracy and convergence rate.
Therefore, the proposed distributed pose graph optimization algorithm can not only reduce the communication overhead in the network of multi-robot systems but also improve the quality of the optimization solution.

\noindent \textbf{Contribution.} This letter is devoted to solving the distributed pose graph optimization problem by multi-robot collaboration.
The main contributions are summarized as follows:
\begin{itemize}
 \item We implement the multi-level graph partitioning method to construct the balanced optimization subproblems in the distributed pose-graph optimization.
 \item Inspired by ACDM, we propose the IRBCD algorithm and prove that the algorithm converges to the first-order stationary point with global optimality.
\end{itemize}

\noindent\textbf{Notations.} 
We define $Sym(h)$ and $Sym(h)^+ $ as the set of $h\times h$ symmetric real and symmetric semi-positive definite real matrices, and denote $I_h$ as the identity matrix of size $h$.
For a matrix $X$, we use $X_{(i,j)}$ to index its $(i,j)$-th entry.
Let $\operatorname{det}(X)$ and $\operatorname{tr}(X)$ be the determinant and trace of a square matrix $X$, respectively.
Given two matrices $A\in\mathbb{R}^{m\times n}$ and $B\in\mathbb{R}^{m\times n}$, we endow the standard inner product with $\langle{A,B}\rangle = tr(A^TB) $.
The notation $\|{\cdot}\|_F $ denotes the frobenius norm of a matrix.
We also define $[n]=\{1,2,\cdots,n\}$.

Matrix manifolds that appear frequently in this letter include the special orthogonal group $SO(d)\coloneqq \left\{R \in \mathbb{R}^{d\times d} \mid R^{T} R=I_{d}, \operatorname{det}(R)=1\right\}$, the special Euclidean group $SE(d)\coloneqq \{(R,T)\in \mathbb{R}^{d\times (d+1)}\mid R\in SO(d), T\in\mathbb{R}^d \}$, and the stiefel manifold $St(r,d)\coloneqq\{R \in {\mathbb{R}^{r \times d}}| R^TR=I_d\}$.

\noindent\textbf{Outline of paper.} The rest of this letter is organized as follows.
Section \uppercase\expandafter{\romannumeral2} reviews the literature related to this work.
Then, Section \uppercase\expandafter{\romannumeral3} gives a mathematical description of the DPGO problem and presents the overall framework of the algorithm.
Finally, the experimental evaluation are carried out in Section \uppercase\expandafter{\romannumeral4}, and the concluding remarks are provided in Section \uppercase\expandafter{\romannumeral5}.

\section{Related Works}\label{sect:related}
Although much research has been done in the SLAM community, this section only reviews the most relevant studies to our work: Pose Graph Optimization and Graph Partitioning.
\subsection{Pose Graph Optimization}
With the development of multi-agent systems, collaborative SLAM has become a practical scheme.
Depending on whether a central coordinator is present or not, the pose graph optimization problem in the backend of the CSLAM system can be described as either a centralized PGO or a distributed PGO.
In centralized PGO, a state-of-the-art method with superior performance is the SE-Sync method with the verifiable correctness guarantee proposed by Rosen et al. \cite{rosen2019se}, \cite{juric2021comparison}.
This core idea has also been applied to other work, such as the description of a low-rank optimization model for global rotational synchronization \cite{chen2021hybrid}.
Furthermore, Briales et al. \cite{briales2017cartan} streamlined the $SE(d)$ synchronization model and extended the search space of the state to the Cartan motion group, achieving faster convergence in specific scenarios.

The distributed PGO problem has also been studied.
For example, Choudhary et al. \cite{7487736} proposed a two-stage Gauss-Seidel method, which iteratively recovers the poses by decoupling the rotation and translation components, but this way is greatly affected by the quality of initialization and cannot guarantee convergence to the first-order critical point.
To address this issue, Fan et al. \cite{fan2020majorization} adopted a \textit{Majorization-Minimization} (MM) method to solve the distributed PGO problem, thus ensuring that its descent sequence could converge to the first-order critical point.
Tian et al. \cite{tian2020asynchronous} also designed an asynchronous parallel distributed algorithm based on Riemannian gradient optimization and analyzed its convergence under limited communication delay.
Inspired by SE-Sync, they further proposed a verifiable distributed pose graph optimization approach based on the \textit{Riemannian block coordinate descent} (RBCD) and proved that it can converge to the global first-order critical point under certain conditions \cite{tian2021distributed}.
There are also some works that consider the robot’s dynamic model and design distributed consensus algorithms based on feedback control theory, such as the GEOD algorithm proposed by Cristofalo et al. \cite{cristofalo2020geod}.
In practical applications, \cite{9686955} put forward Kimera-Multi, a distributed multi-robot cooperative SLAM system, which can robustly identify and reject false loop closure detections under limited communication bandwidth, and construct a globally consistent 3D semantic grid model in real-time.
However, it should be emphasized that all the above methods are based on the local pose information of each robot to perform sequential partitioning operations, without fully considering the structural characteristics of the pose graph.
\subsection{Graph Parititioning}
The graph segmentation problem is traditional and difficult in the combinatorial optimization field.
For example, the minimum-maximum balanced graph segmentation problem belongs to the class of NP-hard problems \cite{sanders2013think}.

Spectral decomposition \cite{cour2005spectral} and maximum-flow minimum-cut \cite{boykov2004experimental} are two frequently used approaches to address the graph partitioning issue.
However, these methods either have high computational complexity or ignore the balance constraint condition, so they are not suitable for dealing with large-scale graph segmentation problems.
In contrast, heuristic, multi-level, and evolutionary algorithms \cite{kuccukpetek2005multilevel}, \cite{ccatalyurek2023more} are relatively sophisticated methods, among which multi-level graph partitioning algorithm has obvious advantages in computation speed and segmentation quality \cite{sanders2011engineering}.

Xu et al. \cite{xu2021bdpgo} introduced an online streaming graph partitioning method for distributed PGO and developed a balanced distributed graph optimization (BDPGO) framework.
However, BDPGO is highly sensitive to dynamic changes in the graph, and its partitioning effects typically depend on the input order of graph data.
\begin{figure}[t]
    \centering
      \centering
    \includegraphics[width=0.5\textwidth]{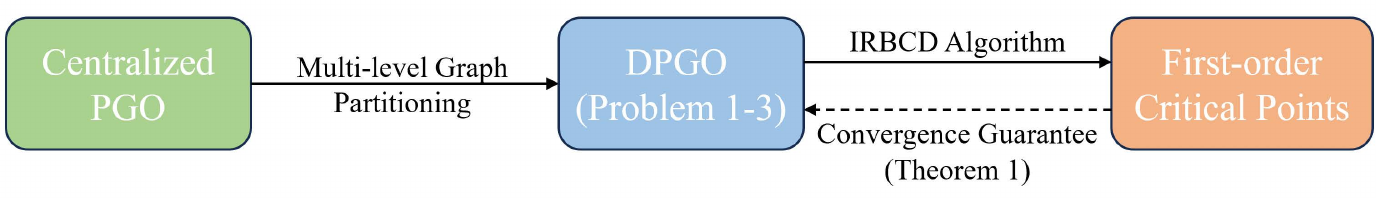}
    \captionsetup{font={small}}
    \caption{
     The relationship among the “Centralized PGO”, “DPGO (Problem 1-3)”, “IRBCD Algorithm”, and “Theorem 1” that are discussed in this work.
     }
    \label{figure3}
\end{figure}

\section{Formulation and Algorithm Design}
In this section, we propose a novel distributed pose graph optimization algorithm (see the pseudocode in Algorithm \ref{algorithm1}).
Firstly, we consider establishing a mathematical model for the distributed pose graph optimization problem after performing multi-level graph partitioning operations.
Following appropriate semi-definite relaxation and \textit{Burer-Monteiro} (BM) decomposition on the subproblems, the low-rank SDP problem is then obtained. Finally, an \textit{improved Riemannian distributed block coordinates descent method} (IRBCD) is applied to solve it.
Figure \ref{figure3} illustrates the internal relationship among the centralized PGO problem, DPGO (Problem 1-3), IRBCD algorithm, and Theorem 1 described in this section.
\subsection{DPGO Problem Formulation}
Consider a multi-robot system consisting of $N$ robots, each of which can only perceive local information.
The purpose of PGO is to estimate the motion trajectories of each robot from noisy relative measurements.
To tackle this problem, we formulate it as a directed graph $\mathcal{G}=\left(\mathcal{V},\mathcal{E}\right)$, where each node in $\mathcal{V}=[n]$ refers to the unknown pose $ x_i^\tau=\left(R_i^\tau,T_i^\tau\right)\in SE(d)$ of robot $i\in[N]$ at its $\tau$ keyframe and each edge in $\mathcal{E}\subset\mathcal{V}\times\mathcal{V}$ corresponds to the relative measurements ${\widetilde{x}}_{i_\tau}^{j_s}=\left({\widetilde{R}}_{i_\tau}^{j_s},{\widetilde{T}}_{i_\tau}^{j_s}\right)\in SE(d)$ between keyframe $\tau$ of robot $i\in[N]$ and keyframe $s$ of robot $j\in[N]$.
Without losing generality, $\mathcal{G}$ is assumed to be weakly connected.
Assume that the measurement model satisfies the following relations:
\begin{equation}\label{40}
\widetilde{R}_{i_\tau}^{j_s}=(R_i^\tau)^TR_j^sR_\epsilon,
\end{equation}
\begin{equation}\label{41}
{\widetilde{T}}_{i_\tau}^{j_s}=(R_i^\tau)^T(T_i^\tau-T_j^s)+T_\epsilon,
\end{equation}
where rotation noise $R_\epsilon$ and translation noise $T_\epsilon$ obey the isotropic Langevin distribution $Langevin(I_d,w_{R})$ and the multivariate Gaussian distribution $\mathcal{N}(0,w_{T})$, respectively \cite{rosen2019se}.

Under the measurement models (\ref{40})-(\ref{41}), we formulate a centralized PGO problem through maximum likelihood estimation:
\begin{equation}\label{60}
\begin{aligned}
\mathop{\min}\limits_{x_i^\tau,x_j^{s} \in \mathcal{V}}&\sum_{\left(x_i^\tau, x_j^{s}\right) \in \mathcal{E}} w_{R}^2\left\|R_{j}^{s}-R_{i}^{\tau} \tilde{R}_{i_{\tau}}^{j_{s}}\right\|_{F}^{2}\\
&+ w_{T}^2\left\|T_{j}^{s}-T_{i}^{\tau}-R_{i}^{\tau} \tilde{T}_{i_{\tau}}^{j_{s}}\right\|_{2}^{2}
\end{aligned}
\end{equation}

To facilitate the collaborative resolution of the aforementioned problem by multiple robots, we perform multi-level graph partitioning on graph $\mathcal{G}$ to partition the pose graph into several balanced subgraphs.

\begin{figure}[t]
    \centering
      \centering
    \includegraphics[width=0.5\textwidth]{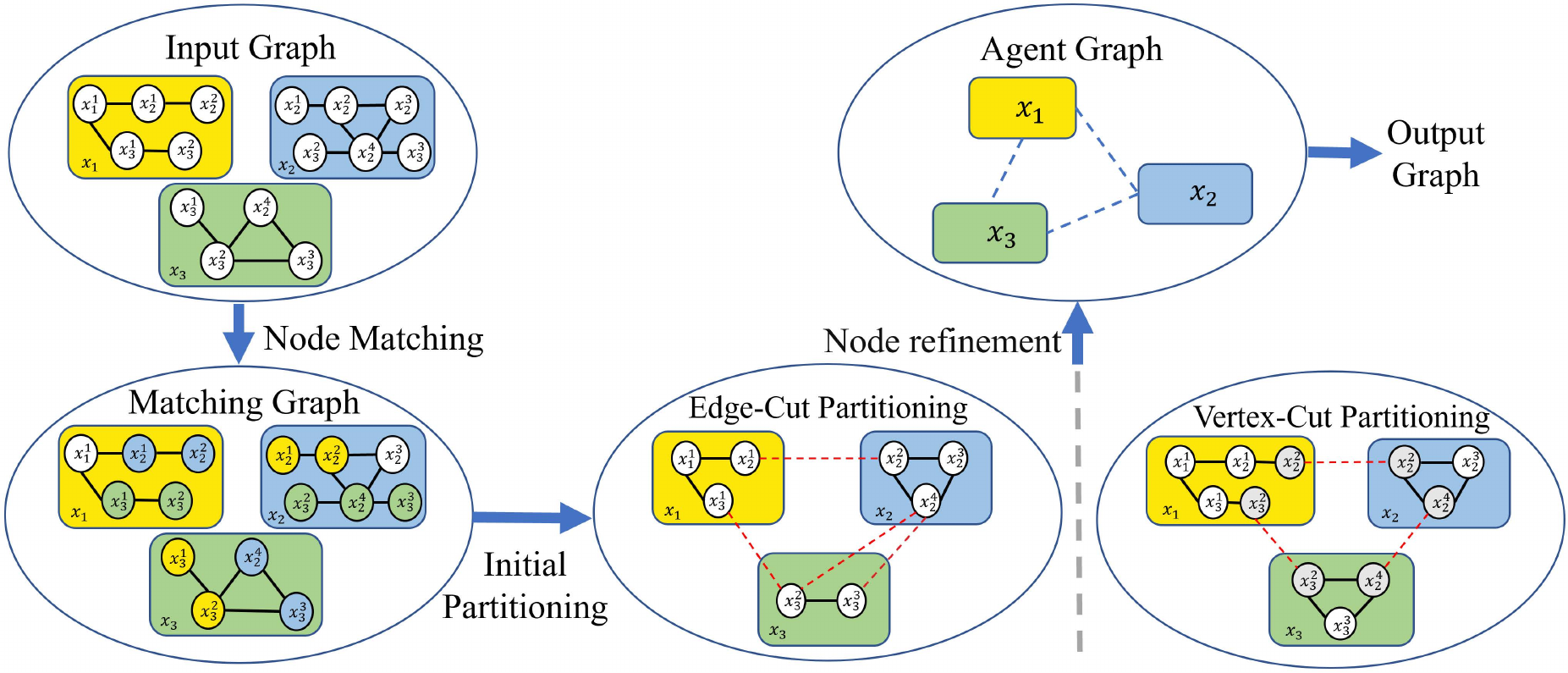}
    \captionsetup{font={small}}
    \caption{The procedure of the multi-level graph partitioning.
         The whole process is divided into three stages: 1) nodes with high coupling degree are matched according to the evaluation function; 2) methods such as edge-cut partitioning or vertex-cut partitioning are used to split the original graph into balanced subgraphs equal to the number of robots; 3) the subgraphs are fine-tuned to generate the agent-level pose graphs.}
    \label{figure1}
\end{figure}

As shown in Figure \ref{figure1}, we select the nodes to be merged according to the edge weight evaluation function in the coarsening stage, then partition the pose graph $\mathcal{G}$  into several balanced subgraphs according to the number of robots by edge-cut or vertex-cut partitioning, and finally refine each subgraph to generate $N$ disjoint sets $X={(x_1,\cdots,x_N)}$.
It is worth mentioning that the state space of each robot $i\in[N]$ not only contains its local pose information but also may extract the pose information from other robots, that is
\begin{equation}\label{42}
{x}_i  = (x_i^1,\cdots,x_j^\tau,\cdots,x_k^{l_i}),\forall i,j,k \in[N], \forall\tau \in[l_i],
\end{equation}
where $[l_i]$ denotes the set of indices of all keyframe poses owned in the $i$-th robot after the graph partitioning.

Therefore, we obtain the following distributed pose graph optimization problem after graph partitioning:

\noindent \textbf{Problem 1(DPGO)}
\begin{subequations}\label{5}
\begin{align}
&~~~\mathop{\min}\limits_{\{x_{i}\}_{i=1}^{N}} f(X)=\sum_{i=1}^{N} f_{i}\left(x_{i}\right)\\
~~f_{i}\left(x_{i}\right)=&~\textstyle\sum_{\left(x_i^\tau, x_j^{s}\right) \in \mathcal{E}_{i}} w_{R}^2\left\|R_{j}^{s}-R_{i}^{\tau} \tilde{R}_{i_{\tau}}^{j_{s}}\right\|_{F}^{2}\nonumber\\
&+ w_{T}^2\left\|T_{j}^{s}-T_{i}^{\tau}-R_{i}^{\tau} \tilde{T}_{i_{\tau}}^{j_{s}}\right\|_{2}^{2} \\
&{x}_i= (x_i^1,\cdots,x_j^\tau,\cdots,x_k^{l_i})\\
\text{s.t.}&~x_i^\tau,x_{j}^{s}\in SE(d),\forall i,j,k \in[N] , \forall \tau, s \in\left[l_i\right],
\end{align}
\end{subequations}
where $f_i$ is the graph optimization subproblem built by each robot, $\mathcal{E}_{i}$ is the edge sets of the local pose graph owned by the $i$-th robot and satisfies $\bigcup_{i=1}^N\mathcal{E}_{i}=\mathcal{E}$.

It has been shown in \cite{briales2017cartan} that the objective function in Problem 1 can be rewritten as $f(X)=\langle G,X^TX\rangle$, where $G \in Sym(n(d+1))$ is a block matrix formed by all relative measurements and weight $(w_{R}^2,w_{T}^2)$.
Let $ Z=X^TX\in Sym((d+1)\times n)^+ $, and then the constraints of Problem 1 are reformulated as:
\begin{subequations}\label{6}
\begin{align}
    Z\succeq0&, \\
    rank(Z)&=d, \\
    Z_{ii(1:d,1:d)}&=I_d, \\
    det(Z_{ii(1:d,1:d)})&=1,
\end{align}
\end{subequations}
where $Z_{ii}$ indicates the $(i, i)$-th block of the matrix $X$.

In order to obtain a better approximate solution or a tighter lower bound for the non-convex optimization problem, we use the convex relaxation method, and replace the non-convex constraint in Problem 1 with its dual problem:

\noindent \textbf{Problem 2(SDP Relaxation for Problem 1)}
\begin{subequations}\label{7}
\begin{align}
  \mathop{\min}\limits_{Z\in Sym((d+1)\times n)^+} f(Z)& = \langle G,Z\rangle \\
  \text{s.t.}~~ Z_{ii(1:d,1:d)}&=I_d, \forall i \in \mathcal{V}.
\end{align}
\end{subequations}

For some $Y \in \mathbb{R}^{r \times (d+1)n}(r\ll n)$, each $Z$ can be decomposed into symmetric low-rank solutions $Z=Y^TY$.
So Problem 2 can be transformed into a convex optimization issue with the low-rank constraints:

\noindent \textbf{Problem 3(Rank restricted problem of P2)}
\begin{subequations}\label{8}
\begin{align}
 &~~~~ \min~~ f(Y)= \langle G,Y^TY\rangle \\
\text {s.t.}~~Y=(Y&_1,T_1,\cdots,Y_n,T_n)\in (St(r,d)\times \mathbb{R})^n,
\end{align}
\end{subequations}
where $Y_i$ belongs to the Stiefel manifold space instead of $SO(d)$, $ \forall i \in\mathcal{V}$.

Ultimately, we can recover the solution $Z^\star$ of SDP relaxation Problem 2 from the first-order critical point $Y^\star$ of Problem 3.
Denote the optimal value of Problem 2 as $f_{\text{SDP}}^\star$.
Considering that the optimal solution of Problem 1 is $f^\star$, and its function value $f(X)$ satisfies the following inequality:
\begin{equation}\label{eq13}
f(X)-f^*\leq f(X)-f_{\text{SDP}}^\star,
\end{equation}
if and only if $f^*=f_{\text{SDP}}^\star$, the equality holds.
Therefore, formula (\ref{eq13}) guarantees the global optimality of the solution yielded from Problem 2 in Problem 1.

\textit{Remark1:} In the execution of graph partitioning tasks within a multi-robot system, there is a central node responsible for receiving all the pose graph structural information instead of raw measurement data.
Utilizing this information, it then proceeds to partition for the rest of the group.
\subsection{Accelerated Riemannian Optimization Algorithm}
Problem 3 is essentially a large-scale optimization problem in the space of product manifold  defined as $\mathcal{M}(r,d)\coloneqq(St(r,d)\times \mathbb{R}^d)^n$, so it's crucial to design an accelerated Riemannian optimization algorithm.

Nesterov has presented an \textit{accelerated coordinate descent} method (ACDM) for the unconstrained large-scale minimization problem which lies in Euclidean space, and its convergence rate is $\mathcal{O}(1/{k^2})$, where $k$ is the number of iterations \cite{doi:10.1137/100802001}.
ACDM adopts a suitable multi-step strategy to update the solution vector sequence $\{Y^k\}\in \mathbb{R}^n$, the estimated solution vector sequence $\{P^k\}\in \mathbb{R}^n$, and the gradient sequence $\{V^k\}\in \mathbb{R}^n$ as follows:
\begin{equation}\label{eq14}
P^k = \alpha_kV^k+(1-\alpha_k)Y^k,
\end{equation}
\begin{equation}\label{eq15}
Y^{k+1} = P^k- \frac{1}{L_{i_k}}\nabla f_i(P^k),
\end{equation}
\begin{equation}\label{eq16}
V^{k+1} = \beta_kV^k+(1-\beta_k)P^k-\frac{\gamma_k}{L_{i_k}}\nabla f_i(P^k),
\end{equation}
where $L_{i_k}$ is the Lipschitz constant of the gradient corresponding to the coordinate block $i_k$ and $\nabla f_i(\cdot)$ is the gradient of the local objective function in Euclidean space.
A sequence of auxiliary variables $\{\alpha_k\},\{\beta_k\},\{\gamma_k\}$ has a coupling equation relationship:
\begin{equation}\label{eq17}
\gamma_k^2-\frac{\gamma_k}{n}= \frac{\beta_k\gamma_k}{n}\frac{1-\alpha_k}{\alpha_k}.
\end{equation}

We extend this acceleration algorithm to the manifold space and call it \textit{Improved Riemannian Block Coordinate Descent} (IRBCD).
In Algorithm \ref{algorithm1}, the ACDM is modified according
to the following four aspects.

1). In order to guarantee that the iteration points generated by IRBCD algorithm are restricted to the manifold space, lines (12) and (16) in Algorithm \ref{algorithm1} introduce an appropriate projection operator ${\rm Proj}_\mathcal{M}\left(\cdot\right)$ to map points $Y^k,P^k,V^k$ in Euclidean space back to $\mathcal{M}(r,d)$.

2). Unlike formula (\ref{eq15}), the block coordinate descending method adopted in this letter generates an iterative sequence $\{Y_{i_k}^{k}\}$ of the chosen coordinate block $i_k\in[N]$ by minimizing the objective function (\ref{8}):
\begin{equation}\label{eq18}
Y_{i_k}^{k+1} = \mathop{\arg\min}_{Y_{i_k} \in \mathcal{M}_{i}} f_{i_k}(P_{i_k}^k),
\end{equation}
where $\mathcal{M}_{i}\coloneqq(St(r,d)\times \mathbb{R}^d)^{l_i}$ denote the search space for the pose of robot $i\in[N]$.

3). Inspired by the generalized momentum acceleration method \cite{siegel2019accelerated}, we utilize the residual between the iterative variables $P^k$ and $Y^{k+1}$ as the momentum term when updating ${V^{k+1}}$, thus avoiding the necessity of calculating the gradient norm involving the Lipschitz constant in the primal method.
To be more specific,
\begin{equation}
V^{k+1} = Proj_{\mathcal{M}}(\beta_{k}V^{k}+(1-\beta_{k})P^k+\gamma_k(Y^{k+1}-P^k)).
\end{equation}
Intuitively, if the target parameters keep moving in the optimal direction, then adding momentum in the same direction can accelerate the algorithm's convergence to a stable point. In addition, the use of momentum information can also help the algorithm gradually get rid of saddle points and avoid stagnation.

4). Our program takes advantage of another type of acceleration mechanism: the adaptive restart strategy. The form of the restart criterion is follows:
\begin{equation}
f(Y^{k+1}>f(Y^k)+c_1\|\text{grad}f(Y^k)\|^2,
\end{equation}
where $c_1$ is the restart constant.
The basic idea is that during the execution of the algorithm, if the value of the objective function does not show a downward trend, the current iteration point is used as the initial value to restart the algorithm to find a new optimal solution, so as to ensure that the IRBCD algorithm can converge on the manifold space.

\textit{Remark2:} Cayley transformation is a common Stiefel manifold retraction.
Assuming that there is $X\in{St(n,p)}$, the expression of the orthogonal projection of $U\in \mathbb{R}^{n\times p}$ on $X$ is
\begin{equation}
{\rm Proj}_X\left(U\right)=U-X\frac{X^TU+U^TX}{2}.
\end{equation}

\textit{Remark3:} The IRBCD technique uses a synchronous updating mechanism to run in a multi-robot network.
The update of sequence $\{P^k\}$(line \ref{P}), $\{V^k\}$(line \ref{V}) just employs local information and projection operator, while the iteration of the target sequence $\{Y^k\}$(line \ref{Y}) requires the shared information $\{P^k\}$ of the robot neighbors.
In addition, the Riemannian gradient information of the objective function for all robots must also be gathered in order to decide whether the algorithm executes the restart operation.
\subsection{Convergence Analysis}
We end this section with the main conclusions related to the convergence of the IRBCD.

\begin{myThe}
\noindent Let $f$ denote the smooth function of the optimization problem (\ref{8}), and suppose the Riemannian gradients of $f$ satisfy a Lipschitz-type condition (see \cite{tian2021distributed}, Lemma 1).
Let $\{Y^k\}$ be generated by Algorithm \ref{algorithm1} with the restart constant $c_1$ and block-specific constants $\lambda$.
Then,
\begin{equation}\label{eq19}
\mathop{\lim}\limits_{k \to \infty} \|\text{grad}f(Y^k)\|=0.
\end{equation}
which implies $Y^k$ is the first-order stationary point as $k\rightarrow\infty$ (see \cite{boumal2020introduction}, Proposition 4.6).
\end{myThe}
\begin{IEEEproof}
To prove the first-order convergence of Algorithm \ref{algorithm1}, we only focus on the update of the sequence $\{Y^k\}$ in IRBCD.
We discuss two cases depending on whether the adaptive restart operation is triggered or not.

Consider the first case when the following occurs if the gap between the function values at time $k$ and time $k+1$ ($k\geq0$) does not satisfy the restart requirements
\begin{equation}\label{eq20}
f(Y^{k})-f(Y^{k+1})\geq c_1\|\text{grad}f(Y^k)\|^2.
\end{equation}
From (\ref{eq20}), $f(Y^k)\geq f(Y^{k+1})$ which means that the global function iterates in a non-increment direction.

For all $M\geq1$, we obtain the following inequality combined with (\ref{eq20}) and a standard telescoping sum argument:
\begin{equation}\label{eq21}
\begin{split}
 f(Y^0)-f(Y^M) \geq c_1 \sum_{k=0}^{M-1}\|\text{grad}f(Y^k)\|^2.
\end{split}
\end{equation}

Assume $f$ is that be globally lower-bound, i.e., there exits $f_{low}\in \mathbb{R}$ such that $f(Y^k)\geq f_{low}$ for all $k$.
Then, taking $M$ as infinity we see that
\begin{equation}\label{eq22}
\sum_{k=0}^{\infty}\|\text{grad}f(Y^k)\|^2 \leq \frac{f(Y^0)-f_{low}}{c_1} \textless +\infty,
\end{equation}
which yields that $\|\text{grad}f(Y^k)\|\rightarrow 0$ as $k\rightarrow\infty$.

In the second case when the restart condition is met.
According to Lemma 2 in \cite{tian2021distributed}, the following bound relationship can be derived:
\begin{equation}\label{eq25}
f(Y^{k})-f(Y^{k+1})\geq \frac{\lambda}{4}\|\text{grad}f(Y^k)\|^2.
\end{equation}
Similar to the proof procedure in the first case, the conclusion (\ref{eq19}) can be proved.
\end{IEEEproof}
\begin{algorithm}[htbp]\small
    \caption{Distributed Pose Graph Optimization Algorithm Procedure with multi-level Partitioning}\label{algorithm1}
    \SetAlgoLined
    \KwIn{Complete Graph $\mathcal{G}$, number of agents $N$,
    Initial solution $Y^0\in\mathcal{M}$, Stopping condition coefficient $\epsilon>0$,
    Restart constant $c_{1}>0$, Initial Auxiliary variables $a_{0},b_{0},\sigma$,Number of iterations $k$}
    \KwOut{First-order critical points $Y^\star$}
    \SetKwProg{Fn}{Function}{}{end}
    \SetKwFunction{GPA}{\textbf{GraphPartitioning Algorithm}}
    \SetKwFunction{FDPGO}{\textbf{IRBCD Algorithm}}
    \tcp{Split the complete graph and obtain the balanced subproblem}
    \Fn{\GPA{$\mathcal{G}$,$N$}}
    {
     Pick Strong/Fast/Eco/Highest algorithm\\
     Partitioning Pose Graph $\mathcal{G}\leftarrow\left(x_1,\cdots,x_N\right)$\\
     Structure the goal function $f:\mathcal{M}(r,d)\rightarrow\mathbb{R}$\\
    }
    \tcp{Riemannian accelerated block gradient method}
    For each agent $i$ solve PGO\\
    \Fn{\FDPGO{$f$,$Y^0$,$\epsilon$,$c_{1}$,$a_{0}$,$b_{0}$,$\sigma$,$k$}}
    {
     $k\leftarrow 0$,$V^0\leftarrow Y^0$

     \While {$\|\text{grad}f(Y^k)\|>\epsilon$}
     {
      \tcp{Update auxiliary variables}
      Compute $\gamma_k\geq \frac{1}{n}$ from $\gamma_k^2-\frac{\gamma_k}{n}=(1-\frac{\gamma_k\sigma}{n})\frac{a_k^2}{b_k^2}$\\
      Let $\alpha_k=\frac{n-\gamma_k\rho}{\gamma_k(n^2-\rho)}$ and $\beta_k=1-\frac{\gamma_k\rho}{n}$\\
      \tcp{Update P}
      $P^k\leftarrow Proj_\mathcal{M}{(\alpha_kV^k+(1-\alpha_k)Y^k)}$\label{P}\\
      Select the next update block $i_k\in[n]$\\
      \tcp{Update Y}
      Updates the selected block:
      $Y_{i_k}^{k+1}\leftarrow\mathop{\arg\min}_{Y_{i_k}\in\mathcal{M}_{i}}f_{i_k}(P_{i_k}^k)$\label{Y}\\
      Leave the values of the other blocks unchanged:
      $Y_{i_{k+1}^{'}}\leftarrow Y_{i_k^{'}}, \forall i_k^{'}\neq i_k$\\
      \tcp{Update V}
      $V^{k+1}=\leftarrow Proj_{\mathcal{M}}(\beta_{k}V^{k}+(1-\beta_{k})P^k+\gamma_k(Y^{k+1}-P^k))$\label{V}\\
      \tcp{Restart Operation}
      \If{$f(Y^k)-f(Y^{k+1})<c_1\|\text{grad}f(Y^k)\|^2$}
      {
      $Y_{i_k}^{k+1}\leftarrow\mathop{\arg\min}_{Y_{i_k}\in\mathcal{M}_{i_k}}f_{i_k}(P_{i_k}^k)$\\
      Leave the values of the other blocks unchanged:
      $Y_{i_{k+1}^{'}}\leftarrow Y_{i_k^{'}}, \forall i_k^{'}\neq i_k$\\
      $V^{k+1}\leftarrow Y^{k+1}$\\
      }
      $b_{k+1} = \frac{b_k}{\sqrt{\beta_k}},a_{k+1}=\gamma_k\beta_{k+1}$\\
      $k\leftarrow k+1$
     }
    }
\end{algorithm}
\section{Experimental Results and Performance Comparisons}
Consider a connected directed network composed of $N=5$ robots and use multithreading to simulate the process of multi-robot collaborative optimization, where each thread is responsible for the iterative optimization of a part of the variables.
The corresponding experiments are run on a workstation: 11th Gen Intel(R) Core(TM) i7-11700 CPU (2.50GHz, 8 cores) and 16GB memory, the operating system is Ubuntu20.04.
All tested algorithms are implemented in C++ and ROPTLIB, where ROPTLIB is a C++ optimization library dedicated to Riemannian optimization \cite{huang2018roptlib}.

The datasets we use are essentially identical to \cite{briales2017cartan}, \cite{tian2021distributed}, encompassing both synthetic and empirical datasets. The synthetic datasets comprise standard evaluation datasets, such as \texttt{Torus}, \texttt{Sphere} as well as \texttt{Manhattan}.
In the real-world datasets, \texttt{Rim} and \texttt{Garage} are the 3D datasets, while \texttt{City}, \texttt{Intel} and others are the 2D datasets.
\subsection{Comparison of multi-level graph partitioning schemes}


The KaHIP \footnote{See https://github.com/KaHIP/KaHIP.} provides four "good" choices of the multi-level graph partitioning algorithm configurations, named \texttt{Strong}, \texttt{Eco}, \texttt{Fast}, and \texttt{Highest}.
Among them, the \texttt{Strong} setting aims at robust partitioning results, the \texttt{Eco} setting targets a trade-off between speed and partitioning quality, the \texttt{Fast} setting strives to reduce execution time, and the \texttt{Highest} setting pursues parallel computing to improve performance \cite{meyerhenke2017parallel}.
We need to select one of these four algorithms that is most appropriate for partitioning pose graphs.

First, we run the four multi-level graph partitioning algorithms on the benchmark datasets and partition the complete pose graph into five pose subgraphs according to the number of robots.
Table \ref{tab:parititions} presents the total number of cut edges between the pose subgraphs under different partitioning methods for 12 datasets.
The experimental results demonstrate that the \texttt{Highest} setting has the best graph partitioning performance, and can decrease the number of cut edges to less than one percent while maintaining that the sizes of the subgraphs are approximately equal.

\begin{figure*}[htbp]
    \centering
	\begin{subfigure}{0.49\linewidth}
		\centering
        \includegraphics[width=0.9\linewidth]{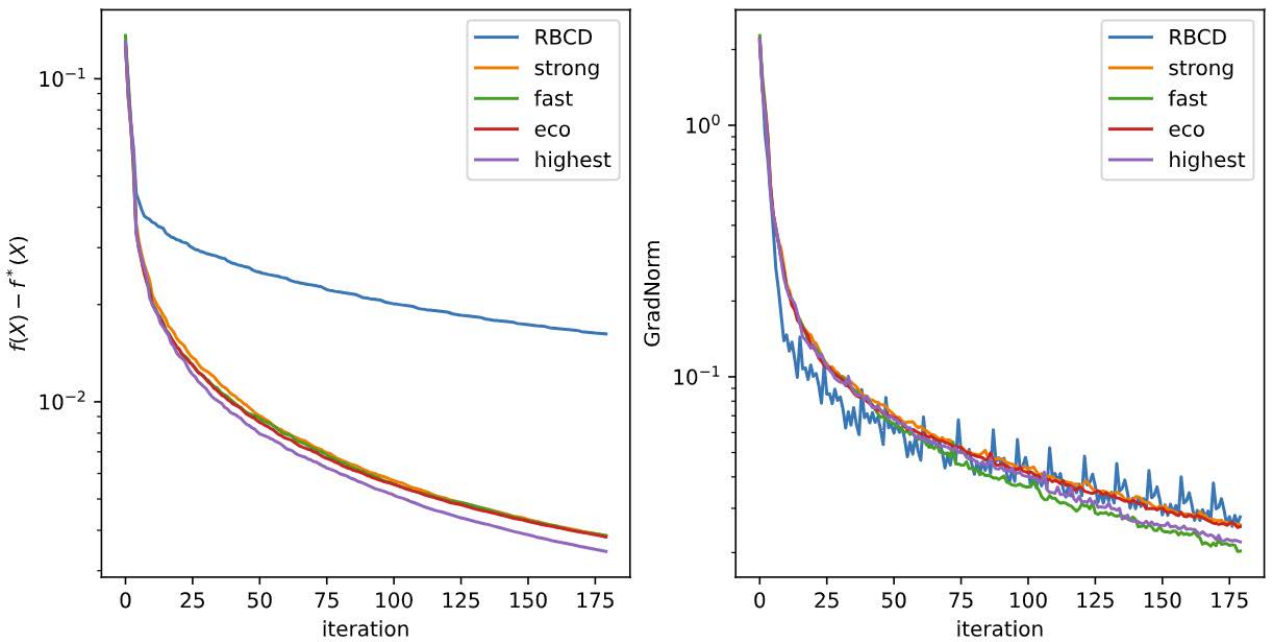}
		\caption{Garage Dataset}
		\label{fig8}
	\end{subfigure}
	\centering
		\begin{subfigure}{0.49\linewidth}
		\centering
        \includegraphics[width=0.87\linewidth]{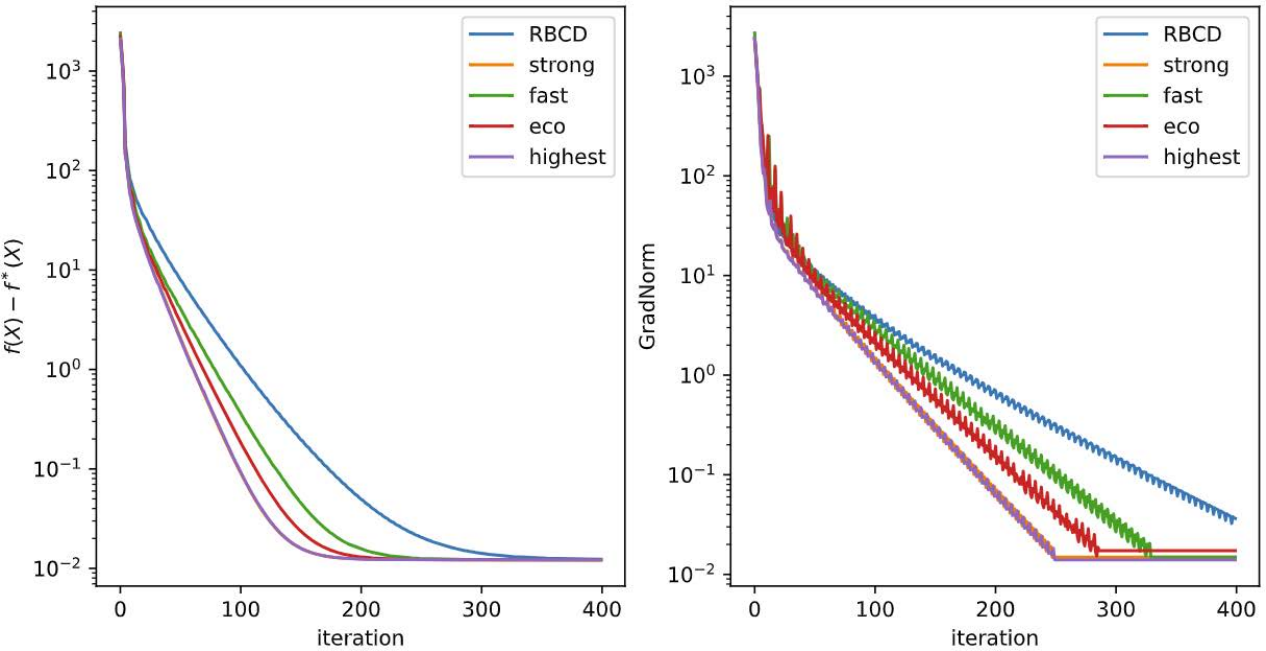}
		\caption{Grid Dataset}
		\label{fig8}
	\end{subfigure}
	\centering
  \captionsetup{font={small}}
  \caption{
  The impact of different graph partitioning methods on the performance of optimization algorithm.
  We integrate each of the four multi-level graph partitioning methods with the RBCD algorithm and conduct experiments on the Garage and Grid datasets.
  Meanwhile, we take the RBCD algorithm based on sequential graph partitioning as the baseline.}
\label{fig30}
\end{figure*}

Then, we adopt the results of different partitioning methods as the block partitioning basis for RBCD, and further explore the impact of different partitioning methods on the algorithm performance.
In this letter, the Riemannian gradient norm \textit{GradNorm} and the optimal gap $f(x)-f^\star$ are two evaluation criteria for the experiment, where $f^\star$ is the optimal value of the centralized SDP relaxation computed using SE-Sync \cite{rosen2019se}.
Figure \ref{fig30} illustrates the iterative process of RBCD under different graph partitioning methods, and compares them with RBCD based on the sequential partitioning algorithm.
The optimization results show that, under the premise of ensuring the convergence of the Riemannian block coordinate descent method, all four partitioning schemes can accelerate the convergence of RBCD, and the algorithm based on the \texttt{Highest} setting has the best performance compared with the other three schemes.

\begin{table}[htbp]
    \centering
    \captionsetup{font={small}}
    \caption{
    \textbf{Performance Comparison of the Partitioning Methods.}
     We apply the baseline method and four multi-level graph partitioning strategies to 12 benchmark datasets and count the number of cut edges between individual variable blocks.
     The baseline method is sequential partitioning and the four graph segmentation strategies are Strong, Eco, Fast, and Highest, respectively.
     }
    \label{tab:parititions}
    \setlength\tabcolsep{3pt}
    \centering
    \small
	\begin{tabular}{ccccccc}
        \toprule
		Datasets & Original Edges & Baseline & Strong & Eco & Fast & Highest  \\
        \midrule
		Rim & 29743 & 2964 & 348 & 302 & 288 & \textbf{243} \\
        City & 20688 & 16724 & 201 & 156 & 135 & \textbf{129} \\
        Csail & 1171 & 232 & 13 & 11 & 10 & \textbf{9}\\
        Intel & 1483 & 444 & 43 & 39 & 40 & \textbf{34} \\
        Grid & 22236 & 3092 & 1184 & 980 &933& \textbf{925}\\
        Torus& 9048 & 798 & 284 & 261 & 252 & \textbf{240} \\
        Sphere  & 4949 & 408 & 182 & 180 & 187 & \textbf{179} \\
        Garage & 6275 & 3715 & 74 & 82 & 61 & \textbf{28}\\
        Cubicle & 16869 & 3276 & 202 & 184 & 180 & \textbf{166}\\
        Kitti02 & 4704 & 94 & 9 & 7 & \textbf{6} & \textbf{6} \\
        Manhattan& 5453 & 1059 & 60 & 48 & 46 & \textbf{41} \\
        Ais2klinik & 16727 & 1348 & 26 & 20 & 15 & \textbf{14} \\
        \bottomrule
	\end{tabular}
    \end{table}

\subsection{Evaluations of the IRBCD Algorithm}
\begin{figure*}[htbp]
    \centering
	\begin{subfigure}{0.49\linewidth}
		\centering
        \includegraphics[width=0.9\linewidth]{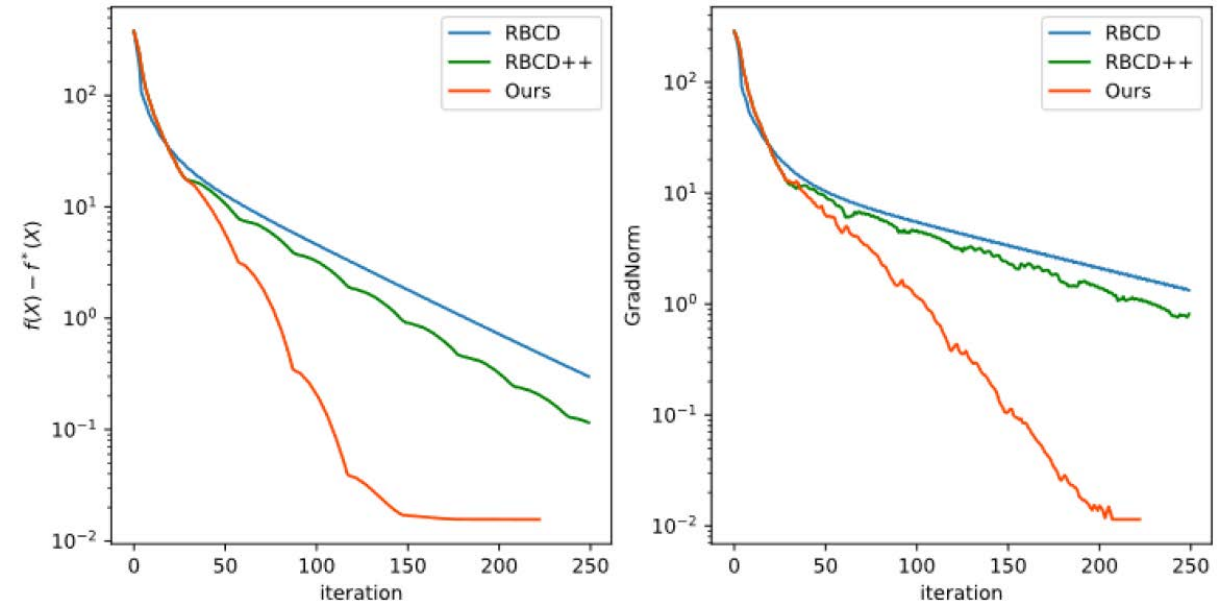}
		\caption{Torus Dataset}
		\label{fig8}
	\end{subfigure}
	\centering
		\begin{subfigure}{0.49\linewidth}
		\centering
        \includegraphics[width=0.9\linewidth]{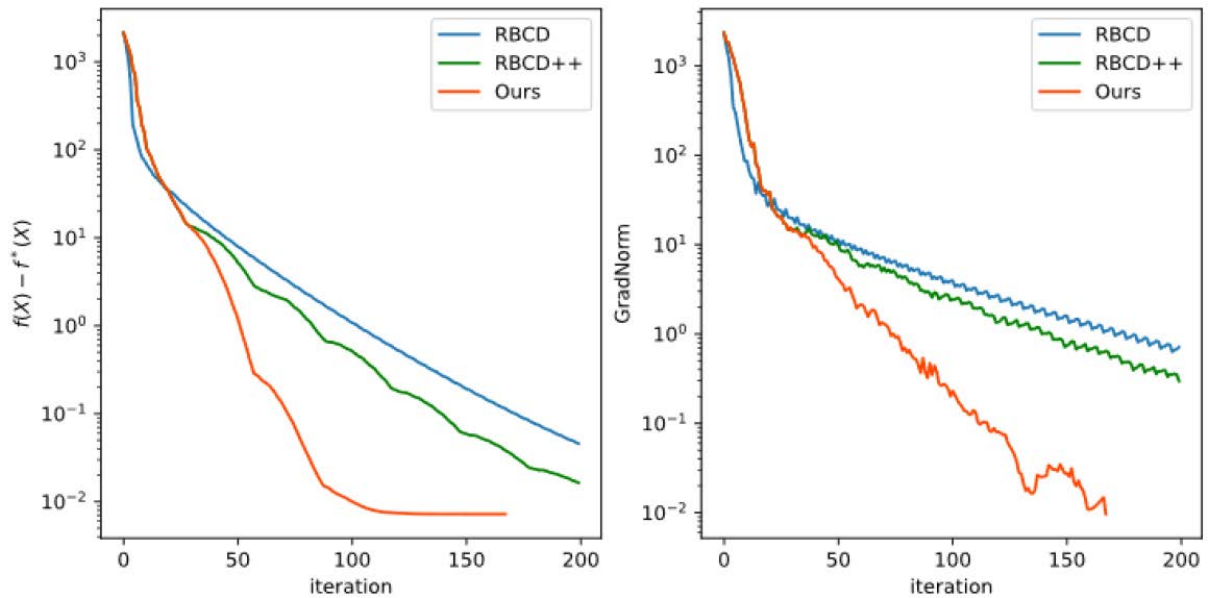}
		\caption{Grid Dataset}
		\label{fig8}
	\end{subfigure}
	\centering
  \captionsetup{font={small}}
  \caption{
  The performance comparison of IRBCD algorithm (ours) and RBCD, RBCD++ optimization algorithms.
  \textcolor{blue}{The Torus and Grid datasets} are used to test the performance of three distributed optimization algorithms, and their convergence results are obtained.}
\label{fig32}
\end{figure*}

We proceed to compare the optimization performance discrepancies between IRBCD (ours), RBCD and RBCD++.
The prior parameter values of the IRBCD method are set as follows: $\sigma=0.001,b_0=1,a_0=\frac{1}{n}$, where $n$ denotes the total dimension of variables.
Figure \ref{fig32} depicts the iterative process of the three algorithms on the Tours and Grids datasets.

The experimental results clearly indicate that, compared with RBCD and RBCD++, the IRBCD algorithm has almost the same computational complexity per iteration, but can converge to the optimal solution with fewer iterations.
For the Grid and Torus datasets, our method only requires less than \textbf{150} iterations to meet the termination criterion $\|\text{grad}f(Y^k)\|\leq0.01$.
From the perspective of the gradient norm change curve, our way has a smaller gradient norm, which indirectly indicates that the obtained solution is closer to the first-order critical point.
Therefore, our proposed method is superior to the current advanced distributed graph optimization algorithms in terms of convergence speed and solution quality.

\subsection{Experimental results of the complete algorithm}
In this part, we further investigate the performance of the distributed pose graph optimization algorithm which combines the \texttt{Highest} graph partitioning and the IRBCD algorithm from multiple dimensions.


To measure the relationship between communication volume and distributed pose graph, we use the communication volume factor as an indicator to reflect the relative data volume that each node needs to transmit in the optimization process, which can be quantitatively expressed as:
\begin{equation}\label{50}
Cvolume=\frac{\sum_{i\in[N] }\sum\nolimits_{v\in x_i}|D(v)|}{|\mathcal{V}|}
\end{equation}
where $|D(v)|$ is the number of external partitions in which vertex $v$ has neighbors \cite{hendrickson2000graph} and $|\mathcal{V}|$ denotes the number of nodes in the pose graph $\mathcal{G}$.

We conduct experiments on five datasets to assess the performance of the complete algorithm and compare it with the benchmark algorithms, namely DGS \cite{7487736} and sequential+RBCD \cite{tian2021distributed}.
As can be seen from Table \ref{tab:dgs_benchmark}, the complete algorithm can obtain the minimum final cost value with fewer iterations.
In addition, we employ the average communication factor as a criterion to discover that the multi-level graph partitioning method can diminish the data transmission volume of each node in the optimization process, consequently alleviating the communication overhead.

\begin{figure}
    \centering
    \includegraphics[scale=0.5]{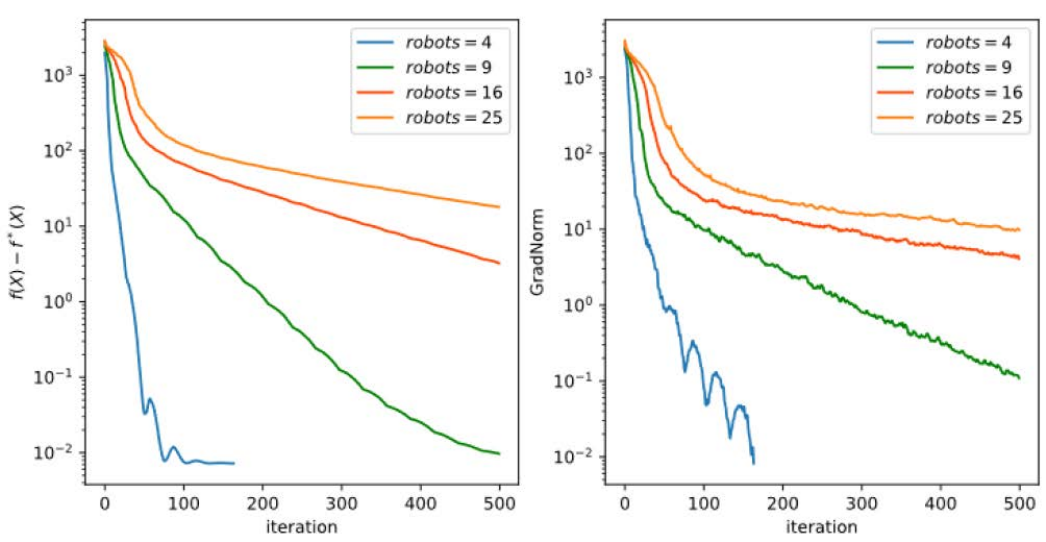}
    \captionsetup{font={small}}
    \caption{Effect of different number of robots on the performance of the complete optimization algorithm.
    The number of virtual robots tested is 4, 9, 16, and 25, respectively.}
    \label{fig31}
\end{figure}

Moreover, we also test the collaborative optimization ability in terms of the number of robots on the Grid dataset.
Figure \ref{fig31} illustrates the impact of robot number changes on algorithm convergence speed.
It can be seen that the complete optimization algorithm can achieve high levels of computational efficiency and solution accuracy when the number of robots does not exceed 16.

\section{Conclusion}

This letter addresses the back-end problem of distributed CSLAM and proposes a novel distributed pose graph optimization algorithm based on multi-level graph partitioning with convergence guarantee.
We first perform multi-level graph partitioning on the original pose graph, so that each robot obtains a subtask with balanced dimensions, and then use the IRBCD method to solve the optimization problem that evolves on the Riemannian manifold space.
Our optimization algorithm has the following merits: on the one hand, it reduces the communication cost of multi-robot parallel computing by the balanced pose graph; on the other hand, IRBCD accelerates the convergence speed of the algorithm.
Compared with existing distributed graph optimization algorithms, our method can converge to a more accurate solution in less time.
We expect to apply our proposed algorithm to practical scenarios in the future to extend its applicability.
Additionally, the iteration graph partition method can be considered to further improve the computational efficiency in online SLAM.

\begin{table*}[]
    \centering
    \captionsetup{font={small}}
    \caption{
    \textbf{Comparison of Proposed Algorithm and Baseline Method.}
    {\textit{Objective Value} is the optimal solution obtained by the different algorithms.
    \textit{Avg.Time} is the average computation time of iteration per robot.
     \textit{ Avg.Iterations} is the average number of iterations per robot. \textit{Avg.Cvolume} is average value of the communication volume factor.
    \textit{ Avg.Partitioning time} is the average time for graph partitioning per keyframe.
}}
    \small
    \label{tab:dgs_benchmark}
    \setlength{\tabcolsep}{1.2mm}{
    \begin{tabular}{c|c|c|c|c|c|c|c}
    \hline
    Dataset   &\begin{tabular}[c]{@{}c@{}}Poses/\\ Edges\end{tabular} & Algorithm      &Objective Value & Avg.Time  & Avg.Iterations        &Avg.Cvolume & Avg.Partitioning time       \\ \hline\hline
    \multirow{3.5}{*}{Sphere(3D)}&\multirow{3.5}{*}{\begin{tabular}[c]{@{}c@{}}2500/\\ 4949\end{tabular}}    & DGS      &1689    &342.7ms          & 60          &    0.38      & -                \\ \cline{3-8}
                                 &
                                 & Sequential+RBCD      &1687   &124.3ms           & 21.6         & 0.33        & 2.3ms                 \\ \cline{3-8}
                                 &
                                & \textbf{Highest+IRBCD} &\textbf{1687}  &\textbf{106.8}ms    & \textbf{18}  &\textbf{0.31} & \textbf{1.8}ms \\ \hline\hline
    \multirow{3.5}{*}{Garage(3D)}  &\multirow{3.5}{*}{\begin{tabular}[c]{@{}c@{}}1661/\\ 6275\end{tabular}}    & DGS          &1.33   &86.2ms            & 5          &    0.15      & -                \\ \cline{3-8}
    &
                               & Sequential+RBCD         &1.31      &72.8ms         & 9.4         & 0.14        & 1.2ms                 \\ \cline{3-8}
                               &
                               & \textbf{Highest+IRBCD} &\textbf{1.26} &\textbf{41.6}ms     & \textbf{4.6}  &\textbf{0.12} & \textbf{0.9}ms \\ \hline\hline
    \multirow{3.5}{*}{Rim(3D)}  &\multirow{3.5}{*}{\begin{tabular}[c]{@{}c@{}}10195/\\ 29743\end{tabular}}  & DGS          &5960.4    &287.4ms           & 103      &    0.42      & -                \\ \cline{3-8}
    &
                               & Sequential+RBCD         &\textbf{5461}    &135.2ms           & 312.6        & 0.34        & 3.5ms                 \\ \cline{3-8}
                               &
                               & \textbf{Highest+IRBCD} &5464 &\textbf{100.4}ms     &\textbf{ 90.8}  &\textbf{0.29} & \textbf{2.7}ms \\ \hline\hline
    \multirow{3.5}{*}{Manhattan (2D)} &\multirow{3.5}{*}{\begin{tabular}[c]{@{}c@{}}5000/\\ 9162\end{tabular}}   & DGS          &242.05   &217.4ms            & 317         &    0.32      & -                \\ \cline{3-8}
    &
                               & Sequential+RBCD         &194.0   &106.4ms          & 157         & 0.21      & 2.6ms                 \\ \cline{3-8}
                               &
                               & \textbf{Highest+IRBCD} &\textbf{193.8}   &\textbf{94.6}ms   & \textbf{139.2}  &\textbf{0.16} & \textbf{2.1}ms \\ \hline\hline
    \multirow{3.5}{*}{City(2D)}  &\multirow{3.5}{*}{\begin{tabular}[c]{@{}c@{}}10000/20687\\ \end{tabular}}  & DGS          &2975.2      &156.3ms        & 493        &    0.27     & -                \\ \cline{3-8}
    &
                               & Sequential+RBCD         &638.7  &108.1ms           & 329.2        & 0.18        &  1.7ms                 \\ \cline{3-8}
                               &
                               & \textbf{Highest+IRBCD} &\textbf{638.6} &\textbf{86.9}ms     & \textbf{304}  &\textbf{0.15} & \textbf{1.3}ms \\ \hline\hline
\end{tabular}}
\end{table*}

\bibliographystyle{IEEEtran}
\bibliography{myRef}

\end{document}